\documentclass[a4paper]{article}

\usepackage{INTERSPEECH2019}
\usepackage{graphicx}
\graphicspath{{./images/}}
\usepackage{latexsym}
\usepackage{url}
\usepackage{amsmath}
\usepackage{booktabs}
\usepackage{colortbl}

\newcommand{\utterance}[1]{\textit{``#1''}}
\newcommand{\phrase}[1]{\textit{`#1'}}
\newcommand{\struct}[1]{\texttt{\small #1}}

\newcommand{\squishlist}{
 \begin{list}{$\bullet$}
  { \setlength{\itemsep}{0pt}
     \setlength{\parsep}{3pt}
     \setlength{\topsep}{3pt}
     \setlength{\partopsep}{0pt}
     \setlength{\leftmargin}{1.5em}
     \setlength{\labelwidth}{1em}
     \setlength{\labelsep}{0.5em} } }

\newcommand{\squishend}{
  \end{list}  }

\title{Neural Named Entity Recognition from Subword Units}
\name{Abdalghani Abujabal$^{1,2}$, Judith Gaspers$^2$}
\address{
  $^1$Max Planck Institute for Informatics\\
  Saarland Informatics Campus, Germany\\
  $^2$Amazon Alexa, Aachen, Germany}
\email{abujabal@mpi-inf.mpg.de, abujabaa@amazon.de, gaspers@amazon.de}

\begin{document}

\maketitle
%

\begin{abstract}
Named entity recognition (NER) is a vital task in spoken language understanding, which aims to identify mentions of named entities in text e.g., from transcribed speech.
Existing neural models for NER rely mostly on dedicated word-level representations, which suffer from two main shortcomings. 
First, the vocabulary size is large, yielding large memory requirements and training time. Second, these models are not able to learn morphological or phonological representations.
To remedy the above shortcomings, we adopt a neural solution based on bidirectional LSTMs and conditional random fields, where we rely on subword units, namely \emph{characters}, \emph{phonemes}, and \emph{bytes}. For each word in an utterance, our model learns a representation from each of the subword units.
We conducted experiments in a real-world large-scale setting for the use case of a voice-controlled device covering four languages with up to $5.5$M utterances per language. Our experiments show that (1) with increasing training data, performance of models trained solely on subword units becomes closer to that of models with dedicated word-level embeddings ($91.35$ vs $93.92$ F1 for English), while using a much smaller vocabulary size ($332$ vs $74$K), (2) subword units enhance models with dedicated word-level embeddings, and (3) combining different subword units improves performance. 
\end{abstract}
\footnotetext[1]{Work done while at Amazon}
\section{Introduction}
\label{sec:introduction}
Named Entity Recognition (NER) is an important task in spoken language technology applications, such as
voice-controlled smart assistants like the Amazon Echo or Google Home. For example, if a user requests an assistant to \utterance{play we are the champions by queen}, an automatic speech recognizer (ASR) can be applied to transcribe the utterance, and subsequently a named entity recogniser can be applied to the ASR output to determine that \phrase{we are the champions} refers to a \struct{song} and \phrase{queen} to an \struct{artist}. As new utterances are collected from the device's users over time, which are annotated with named entities, regular retraining of NER models with increasing data amounts is needed.

Recently, several neural models for named entity recognition have been proposed (e.g., \cite{DBLP:journals/tacl/ChiuN16,DBLP:conf/naacl/LampleBSKD16}), indicating promising performance on rather small and artificially generated datasets~\cite{DBLP:conf/conll/Sang02,DBLP:conf/conll/SangM03}. However, these models either rely on word-level representations or combine them with character-level representations. Such models suffer from the following shortcomings: 
\begin{itemize}
	\item The vocabulary size is large, yielding a large number of parameters, and hence, large memory requirements and training time. This is problematic if large amounts of data are available, in particular if there are memory constraints for the application. 
	\item They ignore the combination with other subword units e.g., phonemes or bytes, which can potentially improve performance by contributing to the modelling of morphology and phonology in case of phonemes. The latter one appears to be particularly useful if a named entity recognizer is applied to transcribed speech. 
	\item Out-of-vocabulary (OOV) words can be problematic. 
\end{itemize}

In this paper, we adopt a neural solution relying solely on subword units, namely \emph{characters}, \emph{phonemes} and \emph{bytes}. 
For each word in an utterance, we learn representations from each of the three subword units. 
The character-level unit looks at the characters of each word, while
the phoneme-level unit treats a word as a sequence of phonemes, using ASR lexica
that map a given word into its corresponding phoneme sequence.
The byte-level unit reads a word as bytes, where we  use the variable length UTF-8 encoding.

A major advantage of subword-based models is the small vocabulary size which can positively affect memory requirements and training time of models. 
This is particularly relevant for large-scale applications and for systems that operate under certain constraints like memory constraints; e.g., handheld or voice-controlled devices.
In addition, subword-units could improve modelling of out-of-vocabulary words and support learning of morphological and phonological information. 
Specifically, character-level networks have already proven to boost the performance of many sequence tagging tasks, including part-of-speech tagging and NER,
in particular for morphologically rich languages~\cite{DBLP:journals/tacl/ChiuN16,DBLP:conf/conll/KleinSNM03,DBLP:conf/naacl/LampleBSKD16}. 
However, while characters have been successfully used to boost NER performance, combining different types of subword units have not yet been explored, leaving open the question if there are additive gains, 
which we address in this paper. 

We present experiments in a large-scale setting for the use case of a voice-controlled device covering four languages, including English, German, and French, with up to 5.5M utterances per language. Our experiments show that: 

\begin{itemize}
\item With increasing training data size, performance of models trained solely on subword units becomes closer to that of models with dedicated word-level embeddings ($91.35$ vs $93.92$ F1 for English), however, with much smaller vocabulary size (e.g. $332$ vs $74$K for English). 
\item Subword units enhance models with dedicated word-level embeddings, in particular, for languages with smaller training data sets.
\item Combining the three subword units (byte-, phoneme- and character-level) yields better results than using only one or two of them.
\end{itemize}


The remainder of this paper is organised as follows. Next, we present our neural network model using subword units. Subsequently, we present experimental results. Before concluding, we discuss related work.

\section{Model}
\label{sec:model}
\begin{figure} [t]
  \centering
    \includegraphics[width=\columnwidth]{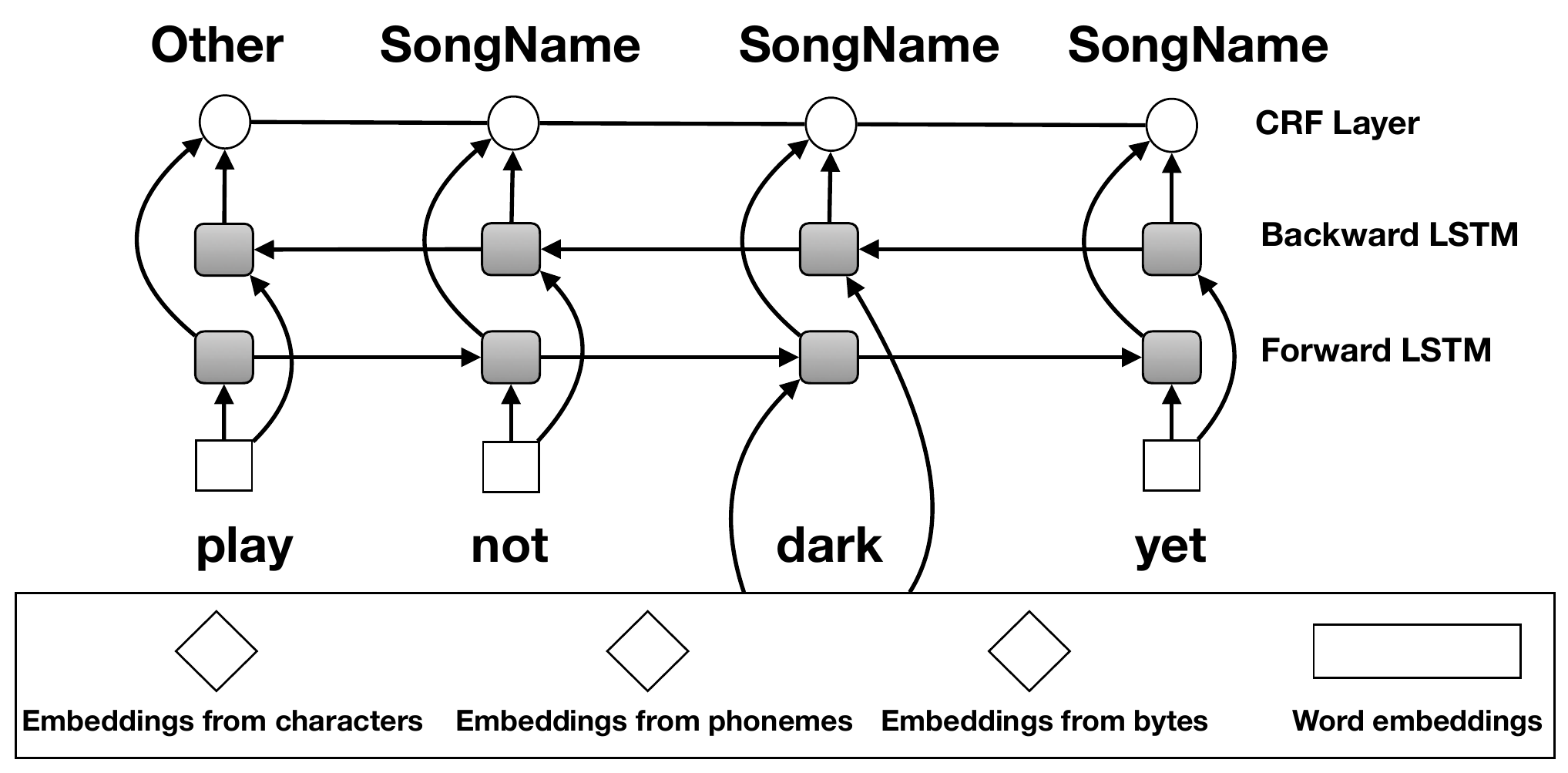}
      \caption{Our model with a word-level bidirectional LSTM layer and CRF layer for decoding. 
      For each word in an utterance, our model learns embeddings from the three subword units. Dedicated word embeddings are optional.}
      \label{fig:model}
\end{figure}

We follow recent work on neural named entity recognition and base our solution on bidirectional LSTMs and conditional random fields (CRF)~\cite{DBLP:journals/tacl/ChiuN16,DBLP:journals/corr/HuangXY15,DBLP:conf/naacl/LampleBSKD16,DBLP:conf/acl/MaH16}.
For each word in an utterance, our model learns a low-dimensional representation from each subword unit (character-, phoneme- and byte-level), which are then concatenated and fed into a bidirectional LSTM-CRF model~\cite{DBLP:journals/corr/HuangXY15,DBLP:conf/naacl/LampleBSKD16}. Our model is depicted in Figure~\ref{fig:model}.
Bidirectional LSTMs capture long-range dependencies among input tokens~\cite{DBLP:journals/nn/GravesS05}.
In this work we use the LSTM implementation that was adopted by Lample et al.~\cite{DBLP:conf/naacl/LampleBSKD16}:
\begin{equation*}
\begin{split}
  i_t = &\sigma(W_{xi}x_t + W_{hi} h_{t-1} + b_i) \\
  c_t = &(1 - i_t) \odot c_{t-1} +  \\
  &i_t \odot tanh(W_{xc}x_t + W_{hc}h_{t-1} + b_c) \\
  o_t = &\sigma(W_{xo}x_t + W_{ho}h_{t-1} + W_{co}c_t + b_o) \\
  h_t = &o_t \odot tanh(c_t), \\
\end{split}
\end{equation*}
where W's are shared weight matrices, b's are the biases, $\sigma$ is the element-wise sigmoid function, 
$x_t$ represents the token at position $t$, $h_t$ is hidden state at time $t$,
and $\odot$ is the element-wise product.
For sequence tagging problems, a softmax layer is used on top of the output of the bidirectional LSTM network to calculate a probability distribution of output tags for a given token.
However, the model assumes independence among output tags, which is not practical for named entity recognition. 
As a remedy, a CRF layer is incorporated for decoding. For details, see \cite{DBLP:conf/naacl/LampleBSKD16}. 

\begin{figure} [t]
  \centering
    \includegraphics[width=\columnwidth]{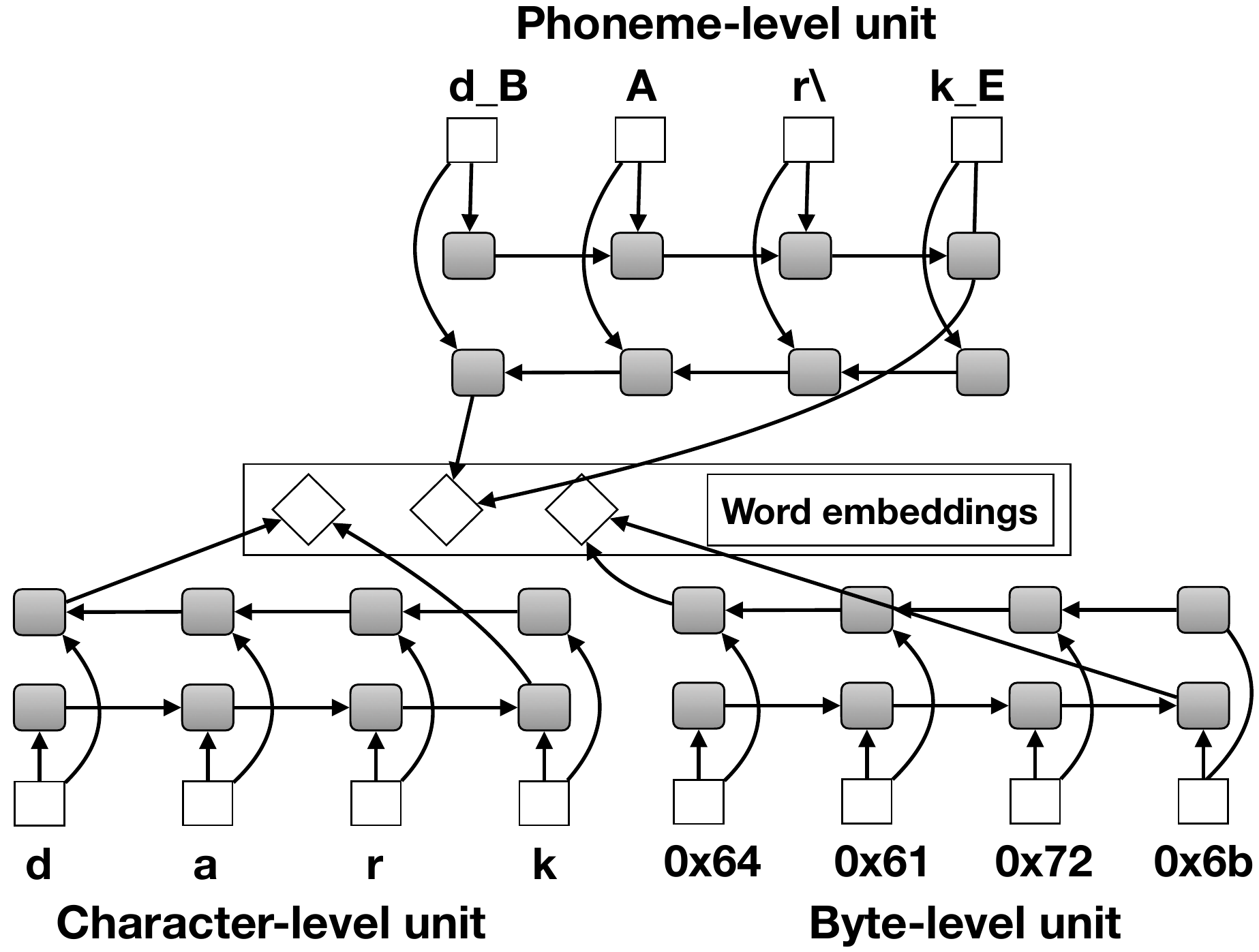}
      \caption{The outputs of the three subword units are concatenated to learn embeddings for the whole word (\phrase{dark}). 
      Optionally, we can concatenate dedicated word embeddings.
      Learned embeddings are then fed into a word-level BiLSTM.  
      }
      \label{fig:subwords}
\end{figure}

{\bf Subword units. } 
We rely on subword units to learn embeddings that represent the full word.
As shown in Figure~\ref{fig:subwords}, each subword unit is a bidirectional LSTM network, where the last hidden state of the forward and the backward networks are concatenated, which constructs $V_c, V_{ph}$ and $V_{by}$ from the character-, phoneme- and byte-level networks, respectively.  The vectors $V_c, V_{ph}$ and $V_{by}$ are, in turn, concatenated to represent the final embeddings of a word. Optionally, we can concatenate dedicated word embeddings that are either randomly initialized or pre-trained.
Subword units enable the model to mitigate the out-of-vocabulary problem, contribute to modelling of morphology and phonology and to have smaller vocabulary size compared to models that rely on word-level representations. 

For the phoneme-level unit, we use lexica that map a given word into its corresponding phoneme sequence. 
Phonemes are represented using the X-SAMPA phoneme set with some additional symbols. For example, the word \phrase{dark} is mapped to $\{d\_B, A, r \setminus, k\_E\}$, where $\_B$ marks the first phoneme in a word, while $\_E$ marks the last one. 
In addition, we add a special symbol $UNK$ to which we map words which are not comprised in our lexica. Including the additional symbols, the final phoneme vocabulary contains $265$ entries with $60$ unique phonemes.
In general, for the setting explored in our paper, i.e. voice-controlled 
devices, phoneme lexica with good coverage are developed for 
the agent'€™s text to speech (TTS) and automated speech recognition components which can be re-used. 
In case lexica with good coverage are not available, tools for grapheme-to-phoneme conversion can be used.

For the byte-level unit, we use the variable length UTF-$8$ encodings to keep the vocabulary small.
For example, \phrase{Sch{\"o}n} is represented as \{0x53 0x63 0x68 0xc3 0xb6 0x6e\}.
Note that the character {\"o} corresponds to two bytes, \{0xc3 0xb6\}. 
This distinguishes this unit from the character-level one.
\section{Experiments}
In the following, we first describe our experimental setup and subsequently present our results.
In our experiments, we explored 1) performance of individual subword units and different combinations of subword units, both with and without using word-level embeddings, 
2) performance of subword only models versus models with dedicated word level embeddings versus models combining both, 
and 3) whether subword units help in the presence of out-of-vocabulary words.
\label{sec:experiments}

\begin{table} [t]
  \centering
    \caption{Number of utterances of per language.}
    \begin{tabular}{l c c c c} \toprule
                               & \textbf{EN}      & \textbf{DE} & \textbf{FR}  & \textbf{ES} \\ \toprule
       Size of  train set          & $3.3$M                & $0.6$M    & $12$K    &  $8$K\\ 
       Size of dev set              & $1.1$M                 & $0.2$M    & $4$K & $2.6$K\\ 
       Size of test set              & $1.1$M                 & $0.2$M    &  $4$K &$2.6$K\\ \bottomrule
    \end{tabular} 

  \label{tab:dataset}
\end{table} 

\subsection{Experimental Setup}
{\bf Datasets. }We use a large real-world dataset covering four languages, namely American English (EN), German (DE), French (FR) and Spanish (ES). 
The data is representative of user requests to voice-controlled devices, 
which were manually transcribed and annotated with named entities. 
Overall, the data covers several domains, comprising different intents and types of named entities.
On average, we have $36$ types of named entities per language. Table~\ref{tab:dataset} shows data statistics for each language. 

{\bf Metric. }To evaluate our models, we use the CoNLL script~\cite{DBLP:conf/conll/Sang02} to compute precision, recall and F1 scores on a per-token basis. We report the average F1 score.

\begin{table} [t]
  \centering
    \caption{Vocabulary size of subword-based versus word-based models.}
    \begin{tabular}{l c c} \toprule
       \textbf{Lang}       & \textbf{Subwords} &     \textbf{Word-level} \\ \toprule
       EN                   & $332$ & $74$K               \\ 
       DE                    & $225$ & $46$K               \\ 
       FR                     & $148$ & $18$K               \\ 
       ES                     & $120$ & $3.7$K               \\ \bottomrule
    \end{tabular} 

  \label{tab:vocab}
\end{table} 

{\bf Training. } We used a mini-batch Adam optimizer~\cite{DBLP:journals/corr/KingmaB14} with a learning rate of $0.0007$ for all the models presented in this paper.  
We tried different optimizers with different learning rates (e.g., stochastic gradient descent), however, they performed worse than Adam. 
The batch size was set to 1024, 256, 4 and 4 utterances for EN, DE, FR and ES, respectively.
The embedding dimension of the subword units is set to 35, while its counterpart of the word-level network is set to 64 (in case dedicated word-level representations are used).
Both subword and word-level networks have a single layer for the forward and the backward LSTMs
whose dimensions are set to 35 and 128, respectively.
We tried several different values, however, the performance was inferior to the one reported with the above values.
When a given number of epochs is reached (40 epochs), training is terminated. 
The model with the best F1 score on the development set is used to make predictions~\footnote{We extended the model at \url{https://github.com/glample/tagger}}.

We used dropout training~\cite{DBLP:journals/corr/abs-1207-0580}, applying a dropout mask to the final embedding layer just before the input to the word-level bidirectional LSTM,
with dropout rate set to $0.5$.

Table~\ref{tab:vocab} shows the vocabulary sizes of different languages for both subword-based and word-level models, 
highlighting the large differences between subword-based models and models with dedicated word embeddings. In addition,
in terms of model complexity, subword-based models have a much smaller number of parameters e.g., for EN, $74$K $*$ $64 = 4.7$M fewer parameters to fine-tune during training.

\begin{table*} [t]
  \centering
    \caption{F1 scores of the subword-only models with different units being used. The model with the three subword units combined achieved best performance across languages, except for FR.}
  \resizebox{\textwidth}{!}{%
    \begin{tabular}{l c c c c c c c} \toprule
     & \textbf{Char} & \textbf{Phoneme} & \textbf{Byte} & \textbf{Char + Phoneme} & \textbf{Char + Byte} & \textbf{Phoneme + Byte} & \textbf{All}\\ \toprule
       EN                   & $89.63$ & $90.3$   & $89.7$   & $91.15$ & $90.58$                     & $91.1$   & $\boldsymbol{91.35}$  \\ 
       DE                   & $84.94$ & $84.21$ & $84.95$ & $86.81$ & $86.32$                     & $86.76$ & $\boldsymbol{87.37}$              \\ 
       FR                     & $80.57$ & $73.65$ & $80.4$   & $80.1$  & $\boldsymbol{82.44}$ & $82.15$ & $81.05$            \\ 
       ES                     & $67.64$ & $62.4$   & $67.07$ & $69.6$  & $70.33$                      & $68.94$ & $\boldsymbol{71.07}$     \\ \bottomrule
    \end{tabular}
     }

  \label{tab:subwords}

\end{table*} 

\begin{table*} [t]
  \centering
    \caption{F1 scores of the combined models with different units being used.} 
  \resizebox{\textwidth}{!}{%
    \begin{tabular}{l c c c c c c c} \toprule
     & \textbf{Char} & \textbf{Phoneme} & \textbf{Byte} & \textbf{Char + Phoneme} & \textbf{Char + Byte} & \textbf{Phoneme + Byte} & \textbf{All}\\ \toprule
       EN                   & $93.96$ & $93.99$ & $93.99$  & $\boldsymbol{94.02}$  & $93.89$  & $93.97$ & $93.88$  \\ 
       DE                   & $90.17$ & $90.02$ & $90.17$ & $\boldsymbol{90.25}$   & $90$      & $90.19$ &  $90.1$                \\ 
       FR                     & $86.37$ & $86.38$ & $86.49$ & $\boldsymbol{87.45}$   & $85.98$ & $85.86$ &  $86.38$                \\ 
       ES                     & $79$ & $80.03$      & $79.08$ & $79.57$  & $79.1$      & $\boldsymbol{80.23}$ & $78.72$     \\ \bottomrule
    \end{tabular}
     }

  \label{tab:combined}

\end{table*} 
\subsection{Results}
\label{subsec:results}
{\bf Subwords only models. }
Table~\ref{tab:subwords} shows the performance of models that rely solely on subword units.
When used individually, different subword units yield the best results for different languages. 
For example, for English, the best individual subword unit is phoneme (with $0.67$ points in F1 more than character), 
while character-level unit achieved best results for French. Here it must be noted that the phoneme lexicon for French had much lower coverage than the one for English, which explains the low F1 score.

When several subword units are combined, results improve for all languages, and except for French,
the best results are achieved when using all of the subword units combined. 
For French, the best combination is characters and bytes, i.e. without using phonemes, which we again attribute to the low coverage of the French phoneme lexicon. 
To explore further whether these improvements  are indeed due to using several subword units rather than the increased dimensionality of the hidden embedding representation, 
we trained models for the different languages using a single subword unit, however,
with higher embedding and LSTM hidden dimensions. The performance was inferior to that reported in Table~\ref{tab:subwords} (last column), 
indicating that there are indeed additive gains from combining different subword units.

{\bf Combined models. }
Table ~\ref{tab:combined} shows results for combining word-level embeddings and subword units.
As can be seen, in this setting there are again additive gains by using several subword units compared to using only one. 
Depending on the language, phonemes yield the best results in combination with either characters or bytes, indicating that phonemes are useful for named entity recognition,
which, to the best of out knowledge, is first explored in this work. The reason might be that they contribute to modelling phonology and/or morphology, thus improving performance in particular for out-of-vocabulary words. 
For three out of the four languages, the combination of characters, phonemes and word-level embeddings achieved the best results.



\begin{table} [t]
  \centering
    \caption{Comparison of subword only models versus word-level models and models combining word-level and subword units. For combined and subword models, the best combination is given. Numbers correspond to F1 values. }
  \begin{small}
    \begin{tabular}{l c c c} \toprule
       \textbf{Lang}       & \textbf{Subwords} & \textbf{Word-level} & \textbf{Combined} \\ \toprule
       EN                   & $91.35$     &     $93.92$($+2.57$)        & $94.02$($+0.1$) \\ 
       DE                     & $87.37$       &   $90.12$($+2.75$)        & $90.25$($+0.13$) \\ 
       FR                      &  $82.44$       &       $86.87$($+4.43$)  & $87.45$($+0.58$)  \\ 
       ES                      &  $71.07$       &       $79.43$($+8.36$)  & $80.23$($+0.8$)  \\  \bottomrule
    \end{tabular} 

  \label{tab:com}
  \end{small}  
\end{table} 

{\bf Comparison. }
Table~\ref{tab:com} compares the performance of models using only subword units, models using only word-level representations and models that combine both. 
We observe that with increasing training data size, performance of models trained solely on subword units becomes closer to that of models with dedicated word-level embeddings ($91.35$ vs
$93.92$ F1 for EN), however, with smaller vocabulary size ($332$ vs $74$K). The gap in performance 
increases as the size of train data decreases ($71.07$ vs $79.43$ F1 for ES). That is,
with sufficient training data, subword-based models achieve rather similar results to word-level ones. 
Models that use both word-level embeddings and subword units achieve the
best results (Table~\ref{tab:com}, last column), showing that subword units can enhance word-level models. 
As train data decreases, the positive effect of subword units increases 
($+0.1$ F1 point for EN and $+0.8$ F1 point for ES). Notably, besides training size, the languages vary in our experiments. In order to draw concrete conclusions on performance in relation to the size of training data, we also trained different models (subword-only, word-level and combined) using different splits of DE train data ($20\%$, $50\%$ and $70\%$).
The results of this experiment confirm our observation that with increasing train data,
performance of subword-based models approaches that of word-level models. Notably, while performance of subwords-only models for EN is still below that of models making use of dedicated word embeddings, the difference may be small enough to opt for the subwords-only model in case there are system requirements e.g., on memory.   

{\bf Out-of-vocabulary words. }
To explore whether our subword units contribute to improved modelling of out-of-vocabulary words, we ran an experiment with the ES data. 
Out of the ES test set $625$ utterances contain at least one out-of-vocabulary word, with $703$ words in total.
F1 values on these utterances are $44.6$, $50.1$ and $51$ for subwords only, word-level and combined models, respectively,
and are thus following the trends observed in Table~\ref{tab:com}. We also computed F1 scores on the out-of-vocabulary words, where, interestingly,  the subword-based model
outperformed the corresponding word-level model ($34.9$ vs $34.8$), while combined model achieved $37.1$ F1, indicating
that subword units are useful in the presence of out-of-vocabulary words.

\section{Related Work}
\label{sec:related}
Named entity recognition is a widely studied problem, where 
methods have been characterized by the use of CRFs with heavy feature engineering, gazetteers and external knowledge resources~\cite{DBLP:conf/acl/FinkelGM05,DBLP:conf/conll/FlorianIJ003,DBLP:conf/emnlp/KazamaT07a,DBLP:conf/conll/KleinSNM03,DBLP:conf/acl/LinW09,DBLP:conf/emnlp/RadfordCH15,DBLP:conf/conll/RatinovR09,DBLP:conf/conll/0001J03,DBLP:conf/interspeech/ZidouniRG10}.
Ratinov and Roth~\cite{DBLP:conf/conll/RatinovR09} use non-local features and gazetteers extracted from Wikipedia, while 
Kazama and Torisawa~\cite{DBLP:conf/emnlp/KazamaT07a} harness type information of candidate entities.
In our work, we opt for a neural solution without hand-crafted features or external resources.

Recently, the focus has shifted towards adopting neural architectures
for NER~\cite{DBLP:journals/tacl/ChiuN16,DBLP:journals/jmlr/CollobertWBKKK11,DBLP:conf/naacl/GillickBVS16,DBLP:journals/corr/HuangXY15,DBLP:conf/naacl/LampleBSKD16,DBLP:conf/acl/MaH16,DBLP:conf/aclnews/SantosG15,DBLP:journals/corr/YangSC16}.
Huang et al.~\cite{DBLP:journals/corr/HuangXY15} use a word-level bidirectional LSTM-CRF for several sequence tagging problems including POS tagging 
and named entity recognition. They made use of heavy feature engineering to extract character-level features.
Lample et al.~\cite{DBLP:conf/naacl/LampleBSKD16} extend the previous model by using a character-level BiLSTM-based unit, where a word is represented by concatenating word-level embeddings  and embeddings learned from its characters. Chiu and Nichols~\cite{DBLP:journals/tacl/ChiuN16} use a convolutional neural network to learn character-level embeddings and LSTM units on the word level. Santos and Guimara{\~e}s~\cite{DBLP:conf/aclnews/SantosG15} propose the CharWNN network, a similar model to that of Chiu and Nichols~\cite{DBLP:journals/tacl/ChiuN16}. 
Gillick et al.~\cite{DBLP:conf/naacl/GillickBVS16} employ a sequence-to-sequence model with a novel tagging scheme. The model relies on bytes, allowing the joint training on different languages for NER, and eliminating the need for tokenization.
Bharadwaj et al. \cite{DBLP:conf/emnlp/BharadwajMDC16} represent words as sequences of
phonemes, which serve as universal representation across languages to facilitate cross-lingual transfer learning.
Finally, Yang et al.~\cite{DBLP:journals/corr/YangSC16} adopt a similar model to that of Lample et al.~\cite{DBLP:conf/naacl/LampleBSKD16}, however, they replaced LSTMs with Gated Recurrent Units (GRUs). Furthermore, they studied the multi-lingual and multi-task joint training, which we plan to address in the future.\\
Overall, existing neural methods for named entity recognition rely mostly on dedicated word embeddings rather than learning such representations from subword units. 
While some work has also addressed characters or bytes, combining different types of subword units has not been explored, which we address in this work. For a comprehensive survey on NER, see \cite{DBLP:conf/coling/YadavB18}.
\section{Conclusion and Future Work}
\label{sec:conclusion}
We presented a neural model for named entity recognition using three subword units: characters, phonemes and bytes. 
For each word in an utterance, the model learns a representation from each of the three subword units,
which are then concatenated and fed into a word-level bidirectional LSTM and CRF layer for decoding. 
Our experiments show that i) with increasing training data, performance of models trained
solely on subword units becomes closer to that of models with dedicated word embeddings, while using a much
smaller vocabulary and fewer trainable model parameters, ii) subword units enhance models with dedicated word embeddings, and iii) combining subword units improves performance.\\
In this paper, we used ASR lexica to map words into sequences of phonemes. An interesting direction for future work is the application together with an automatic speech recogniser, where phonemes are transcribed from the speech utterances and subsequently used in the NER model. This could improve NER performance further by taking additionally information directly from the speech signal into account, which could be in particular useful for modelling homonyms.

\bibliographystyle{IEEEtran}

\bibliography{ner}


\end{document}